\documentclass[conf]{new-aiaa}
\usepackage[utf8]{inputenc}

\usepackage{graphicx}
\usepackage{amsmath}
\usepackage[version=4]{mhchem}
\usepackage{siunitx}
\usepackage{longtable,tabularx}
\usepackage{algorithm}
\usepackage[noend]{algpseudocode}
\usepackage{algpseudocode}
\usepackage{tabularx}
\usepackage{subfigure}

\title{Reinforcement Learning Based Self-play and State Stacking Techniques for Noisy Air Combat Environment }

\author{Ahmet Semih Tasbas
\footnote{A. S. Tasbas is with the ASELSAN Research Center, Ankara 06370, Turkey, and also with the Department of Computer Engineering, Istanbul Technical University, Maslak, Istanbul 34469, Turkey (contact e-mail:stasbas@aselsan.com.tr and tasbas20@itu.edu.tr).} }
\affil{Istanbul Technical University, Istanbul, Turkey, 34469}

\author{Safa Onur Sahin  \footnote{S. O. Sahin is with the ASELSAN Research Center, Ankara 06370, Turkey, and also with the Department of Electrical and Electronics Engineering, Bilkent University, Bilkent, Ankara 06800, Turkey (contact e-mail:sosahin@aselsan.com.tr).}}
\affil{ASELSAN, Ankara, Turkey, 06200 }

\author{Nazim Kemal Ure \footnote{N. K. Ure is with the Department of Aeronautics and Astronautics, Istanbul Technical University, Maslak, Istanbul 34469, Turkey (contact e-mail: ure@itu.edu.tr).}}
\affil{Istanbul Technical University, Istanbul, Turkey, 34469}

\begin{document}

\maketitle

\begin{abstract}
Reinforcement learning (RL) has recently proven itself as a powerful instrument for solving complex problems and even surpassed human performance in several challenging applications. This signifies that RL algorithms can be used in the autonomous air combat problem, which has been studied for many years. The complexity of air combat arises from aggressive close-range maneuvers and agile enemy behaviors. In addition to these complexities, there may be uncertainties in real-life scenarios due to sensor errors, which prevent estimation of the actual position of the enemy. In this case, autonomous aircraft should be successful even in the noisy environments. In this study, we developed an air combat simulation, which provides noisy observations to the agents, therefore, make the air combat problem even more challenging. Thus, we present a state stacking method for noisy RL environments as a noise reduction technique. In our extensive set of experiments, the proposed method significantly outperforms the baseline algorithms in terms of the winning ratio, where the performance improvement is even more pronounced in the high noise levels. In addition, we incorporate a self-play scheme to our training process by periodically updating the enemy with a frozen copy of the training agent. By this way, the training agent performs air combat simulations to an enemy with smarter strategies, which improves the performance and robustness of the agents. In our simulations, we demonstrate that the self-play scheme provides important performance gains compared to the classical RL training.

\end{abstract}

\section{Nomenclature}

{\renewcommand\arraystretch{1.0}
\noindent\begin{longtable*}{@{}l @{\quad=\quad} l@{}}
$\psi$  & bank angle of aircraft \\
$\Delta_{\psi}$ &  bank angle change  \\
$v$ & velocity \\
$\Delta_{v}$ & velocity change of aircraft \\
$x$ & aircraft position in x-axis \\
$y$ & aircraft position in y-axis \\
$\hat{x}$ & noisy aircraft position in x-axis \\
$\hat{y}$ & noisy aircraft position in y-axis \\
$\hat{\psi}$ & noisy bank angle of aircraft \\
$ATA$ & antenna train angle \\
$AA$ & aspect angle \\
$LOS$ & line of sight \\

\end{longtable*}}

\section{Introduction}
\lettrine{M}{odern} aircraft such as F-35 and F-16 are produced in agile structures that can make sudden maneuvers to avoid enemy aircraft. The military pilots experience a long and tough training process to use these aircraft. Although they show a decent maneuvering performance on these aircraft, their performance is inherently restricted by the human reflexes and G-load tolerance. On the other hand, the autonomous control of the unmanned aircraft (UA) can enable to utilize the full maneuvering capacity of these aircraft and also prevent the possible damages to the human pilots \cite{2010_dynamic_programming}. As stated by Yang et al. \cite{2020_short_range}, the design of an autonomously maneuvering aircraft is a highly challenging problem since these aircraft fight each other with aggressive maneuvers at close range. In addition, unpredictable actions of the enemy aircraft increases the problem complexity. As a remedy, the deep learning methods and the optimization algorithms can be used to tackle this problem.

\par As one of the main branches of the artificial intelligence, the RL techniques demonstrate prominent results in several decision-making problems. For example, trajectory generation in drone racing \cite{2021_drone_racing}, large-scale strategy management in Dota 2 game \cite{2019_dota2_rl}, and autocurricula training in contentious multi-agent environment \cite{2019_emergent_tool_use} are the certain illustrative RL based solutions. Similar to these problems, the air combat problem requires sequentially taking reasonable actions with only a limited feedback from the environment such as the win/loss of the fight, therefore, the RL based solutions are highly suitable for this problem.

The air combat problem is extensively studied in the literature. McGrew et al. \cite{2010_dynamic_programming} present a method based on approximate dynamic programming.  They pointed out that the success of their approach depends on feature development, reward shaping, and trajectory sampling. Isci et al. \cite{2022_Energy_Budgets} emphasize that it is important to protect the energy of the aircraft along with defeating the enemy aircraft in air combat. They define an energy condition for aircraft and penalized the agent in reward function based on the energy consumption. Unlike the other studies, Yang et al. \cite{2020_short_range} define the probability of shooting the enemy aircraft as being directly proportional to the proximity of the enemy. In addition, they also divide the training phase into two parts. In the first part, the enemy takes simple actions. In the second part, the agent plays against a more sophisticated enemy with a stochastic policy. Pope et. al. \cite{2021_hierarchical_air_combat} employ a hierarchical reinforcement learning methodology. They develop their approach on DARPA's AlphaDogFight Trials environment. This simulator provides low-level controller input. Therefore, they separate the control structure into different layers and trained different agents for specific tasks. 

\par In all of these studies, the environment is assumed to be noise-free, i.e., the exact position of the enemy aircraft is known. However, this may not be achievable in a real-life scenario. For example, the sensor errors and medium conditions may cause the observation to differ from the reality. In this study, contrary to previous studies, we first develop a simulation that provides noisy observations to the agents. We also emphasize that the noise characteristics are adjustable through its parameters, hence, we can evaluate the performance of the trained agents under a wide range of noise level. Since the agents fights against the enemy aircraft by using the inaccurate observations, the air combat problem becomes more complex, hence, the performance of the RL algorithms usually decreases as the noise level increases. In our algorithm, we stack the consecutive observations to address deteriorating effect of the corrupted observations on the performance of the RL algorithms.

\par Another issue emphasized in this study is the strategy of the enemy agent. In previous studies, the strategy of the enemy agent was generally chosen as follows: predefined maneuvers \cite{2022_Energy_Budgets,2020_short_range,2021_hierarchical_air_combat,2021_Beyond_Visual_range}, Minimax \cite{2018_Deep_Q_Ma, 2020_Inverse_RL} and heuristic search algorithms (Monte Carlo Tree Search) \cite{2018_Deep_Q_Ma}. Contrary to previous studies, we aim to create a smarter enemy behavior using competitive self-play techniques. Therefore, the agent encounters smarter enemy strategies, and the problem gets harder progressively during training. We create different self-play structures to demonstrate the effect of the enemy difficulty performance of the training agent. Furthermore, we empirically show that the agent trained with self-play scheme outperforms the agent trained without self-play in a noisy test environment. 

\par The rest of the paper is organized as follows. Section III provides information about problem formulation, including aircraft dynamics, air combat geometry, noisy state space generation, and action space. Section IV introduces partially observable Markov decision processes, Deep Q Learning (DQN), a value-based deep reinforcement learning algorithm, reward function design, noise reduction with state stacking, and competitive self-play techniques. Section V provides empirical results of the proposed methods. Finally, section VI concludes the paper with final remarks.

\section{Problem Formulation}

\subsection{Aircraft Dynamics and Air Combat Geometry}
We design and construct a simulator, which models the motion dynamics of the aircraft and performs a close-range one-to-one air combat scenario. Since the deep reinforcement learning based algorithms usually require a large number of interactions with the environment, we assume that the inner loop controllers of the aircraft are well-tuned and use a simplified model for the state of the aircraft. 

\par The state-transition model is inspired by previous studies {\cite{2010_dynamic_programming, 2018_Deep_Q_Ma}} with the following equations:

\begin{equation}
\label{equ_psi_update}
\psi = \psi + a_{\psi} \Delta_{\psi},
\end{equation}

\begin{equation}
\label{equ_velocity_update}
v = \text{clip}( v + a_{v} \Delta_{v}, v_{min}, v_{max} ),
\end{equation}

\begin{equation}
\label{equ_posx_update}
x = x + v_{t} \cos({\psi  \frac{\pi}{180^\circ} }),
\end{equation}

\begin{equation}
\label{equ_posy_update}
y = y + v_{t} \sin({\psi  \frac{\pi}{180^\circ} }).
\end{equation}

In \eqref{equ_psi_update} and \eqref{equ_velocity_update}, the bank angle change $\Delta_{\psi}$ and the velocity change $\Delta_{v}$ selected as $10^{\circ}$ and 0.1 m/s, respectively. We also limit the minimum and maximum velocity of the aircraft as $v_{min}=4 \text{ m}/\text{s}$ and $v_{max}=8 \text{ m}/\text{s}$.

\par
\begin{figure}[hbt!]
    \centering
    \includegraphics[width=0.58\textwidth]{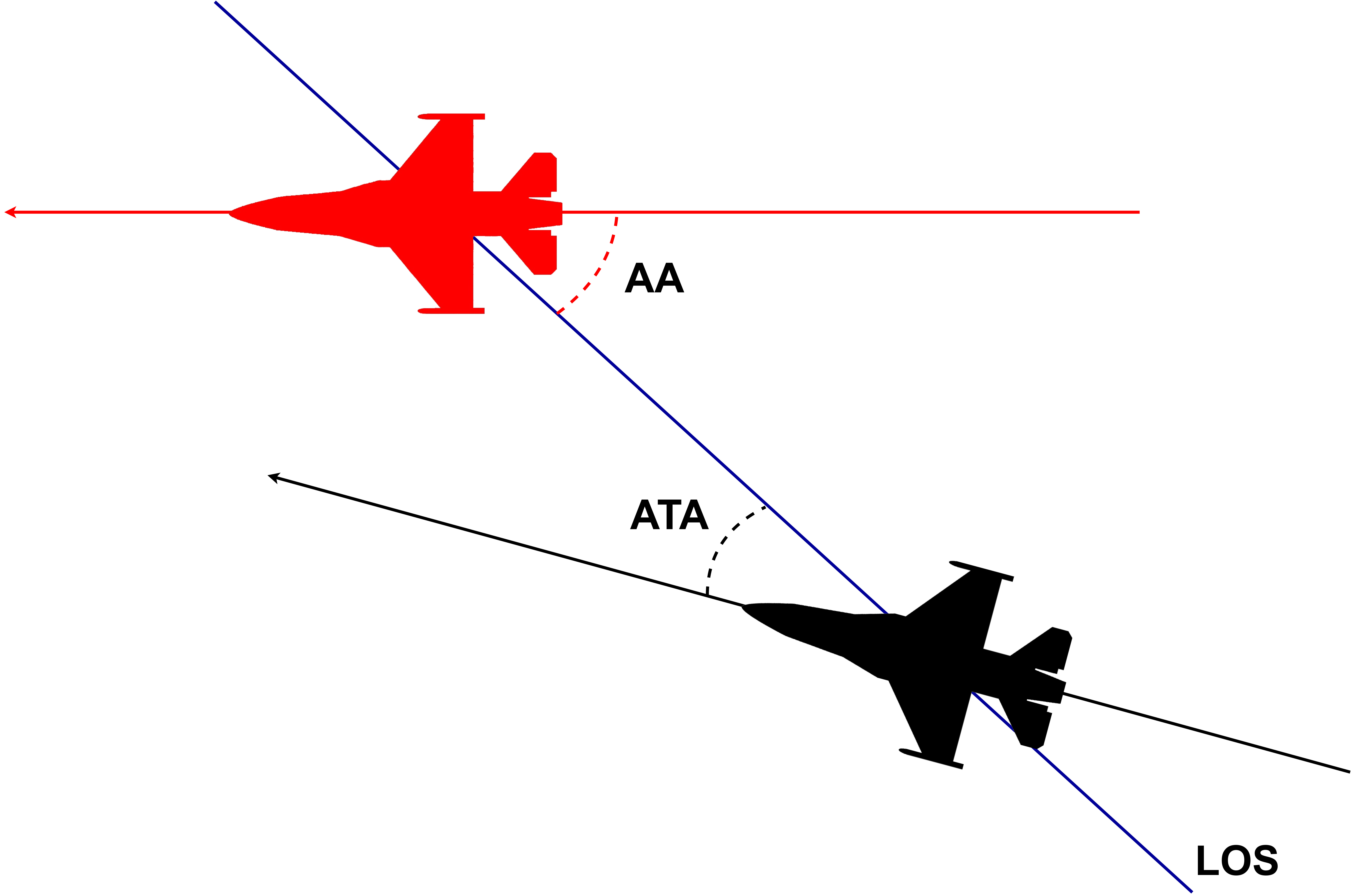}
    \caption{Representation of a close-range one-to-one air combat scenario, where we control the black aircraft and the red aircraft is the enemy. It also illustrates antenna train angle (ATA), aspect angle (AA), and line of sight (LOS).}
    \label{fig:ata_aa}
\end{figure}

 In Fig. \ref{fig:ata_aa}, we describe a close-range air combat scenario to clarify the problem, where the black aircraft is the agent and the red aircraft is the enemy. Here, LOS is the line of sight, ATA is the antenna train angle, and AA is the aspect angle. LOS refers to the line passing through the center of gravity of two aircraft. ATA refers to the angle between the agent's heading and the enemy aircraft's position. AA denotes the angle between the enemy aircraft's tail and LOS. When both ATA and AA values get closer to zero, the agent gets the position, where it can shoot the enemy. 

\subsection{Noisy State Space}
The state transition model  generated from aircraft dynamics is specified in the previous section. It is assumed that the enemy aircraft's position is absolutely correct in these dynamics. However, it is usually impossible to obtain noise-free data in real life. In order to overcome this problem, we suggest to add noise to the state space during the training and testing phases. The noise-free state space in the environment is: 

\begin{equation}
\label{equ_noise-free-state-space}
s = [R, ATA_{a}, ATA_{e}, \psi_{a}, \psi_{e}, v_a, v_e ],
\end{equation}

\noindent where the distance $R$, relative positions of aircraft $ATA_{a}$ and $ATA_{e}$ depend on the position of the aircraft. Applying different noise to each of these parameters may lead to inconsistencies. Hence, we insert the noise into the enemy aircraft's parameters before calculating these values. The details of the noisy state generation are shown in Algorithm \ref{algo:noisy-state-generation}. The noise values are derived from the Gaussian distribution. The position noise $n_{x,y} \sim \mathcal{N}(\mu,\,\sigma^{2}_{x,y})$ is inserted to the enemy aircraft's position $x_{e}$ and $y_{e}$, whereas the angle noise $n_{\psi} \sim \mathcal{N}(\mu,\,\sigma^{2}_{\psi})$ is inserted to the enemy aircraft's angle $\psi_{e}$. 
Since neural networks are used as a function approximator, we normalize state space between 0 and 1 to ensure stability during the training stage. 

\begin{algorithm}
\caption{Noisy State Generation }\label{algo:noisy-state-generation}
\textbf{Input}: agent status = $\left\{{x_{a}, y_{a}, \psi_{a}, v_{a}}\right\}$, enemy status = $\left\{{x_{e}, y_{e}, \psi_{e}, v_{e}}\right\}$, $n_{x}, n_{y} \sim  \mathcal{N}(\mu,\,\sigma^{2}_{x,y})$,  $n_{\psi} \sim \mathcal{N}(\mu,\,\sigma^{2}_{\psi})$.
  \begin{algorithmic}[1]
  
    \State $ \hat{x}_{e} = x_{e} + n_x, \text{where } n_x \sim \mathcal{N}(\mu, \sigma^2_{x,y}) $
    \State $ \hat{y}_{e} = y_{e} +  n_y, \text{where } n_y \sim \mathcal{N}(\mu, \sigma^2_{x,y}) $
    \State $ \hat{\psi}_{e} = \psi_{e} +  n_{\psi}, \text{where } n_{\psi} \sim \mathcal{N}(\mu, \sigma^2_{\psi}) $
    
    \State $ \hat{R}=\sqrt{ (x_{a}-\hat{x}_{e})^2 + (y_{a}-\hat{y}_{e})^2 } $
    \State $\hat{ATA}_{a} = ATA(x_{a}, y_{a}, \hat{x}_{e},\hat{y}_{e}, \psi_{a} )$
    \State $\hat{ATA}_{e} = ATA(\hat{x}_{e},\hat{y}_{e}, x_{a}, y_{a}, \hat{\psi}_{e} )$
  \end{algorithmic} 
\textbf{Output}: $s_{t+1} = [\hat{R},\hat{ATA}_{a}, \hat{ATA}_{e}, \psi_{a}, \hat{\psi}_{e}, v_a, v_e ]$ 
  
\end{algorithm}

\par
\begin{figure}[hbt!]
    \centering
    \subfigure[]{\includegraphics[width=0.49\textwidth]{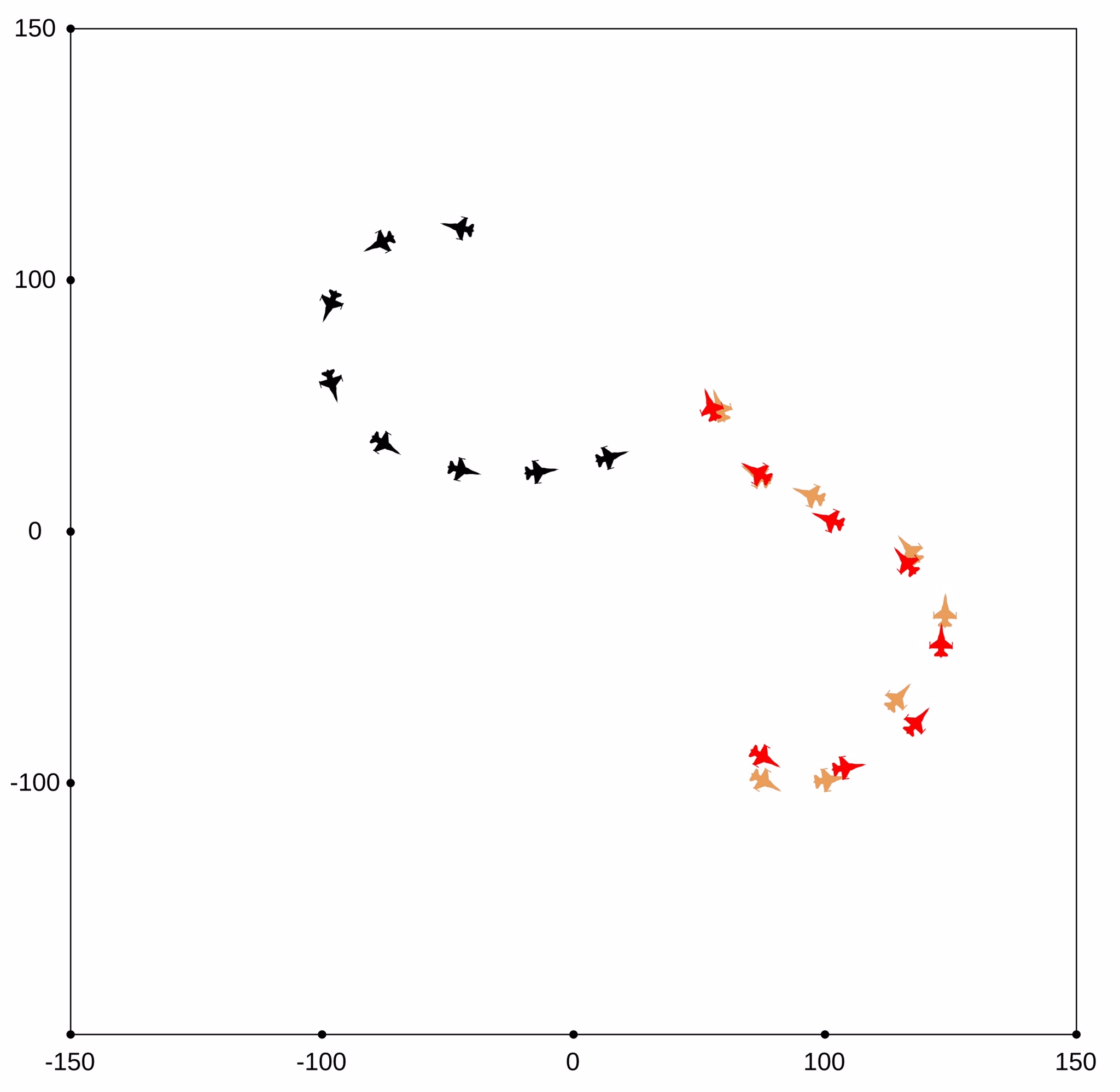}} 
    \subfigure[]{\includegraphics[width=0.49\textwidth]{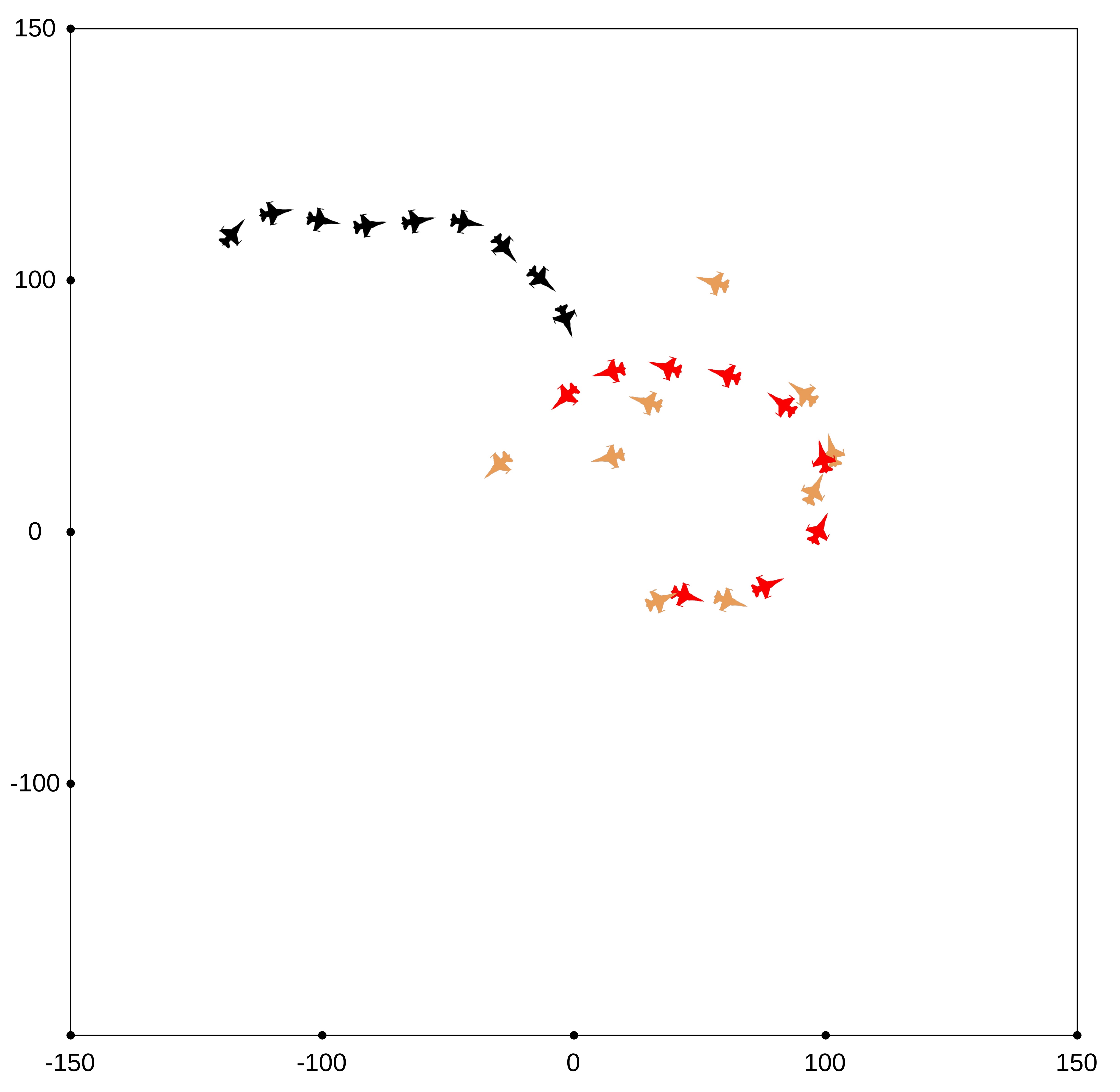}} 
    \caption{ Visualization of the effect of the noisy state space on the environment. The black plane is the training agent, the red plane is the enemy agent, and the orange plane is how the training agent sees the enemy agent. (a) The position noise value is $n_{x,y} \sim \mathcal{N}(0,5)$, and the angle noise value is $n_{\psi} \sim \mathcal{N}(0,1)$ (b) The position noise value is $n_{x,y} \sim \mathcal{N}(0,20)$, and the angle noise value is $n_{\psi} \sim \mathcal{N}(0,1)$.}
    \label{fig:noisy_state}
\end{figure}

 Fig. \ref{fig:noisy_state} demonstrates the actual and the noisy positions of the enemy aircraft. In this figure, the black plane is the training agent, the red plane is the enemy agent, and the orange plane is how the training agent sees the enemy agent. In part (a) of this figure, the enemy's position noise value is $n_{x,y} \sim \mathcal{N}(0,5)$ and the angle noise value is $n_{\psi} \sim \mathcal{N}(0,1)$. In part (b) of this figure, the enemy's position noise value is $n_{x,y} \sim \mathcal{N}(0,20)$ and the angle noise value is $n_{\psi} \sim \mathcal{N}(0,1)$. Since the noise in part (b) is high, the training agent perceives the opponent as more dispersed.

\subsection{Action Space}
Aircraft take two different control inputs as an action; velocity $v$ and angle $\psi$:

\begin{equation}
\label{equ_action_space}
a = [v, \psi].
\end{equation}

These two parameters can be ascent, descend or remain same in discrete space. Therefore, the combination of these two parameters composes nine discrete actions:
\begin{gather}
\begin{bmatrix}
v_{t}-\Delta_{v}, \psi_{t}+\Delta_{\psi} && v_{t}, \psi_{t}+\Delta_{\psi} && v_{t}+\Delta_{v},  \psi_{t}+\Delta_{\psi} \\
v_{t}-\Delta_{v}, \psi_{t} && v_{t}, \psi_{t} && v_{t}+\Delta_{v}, \psi_{t} \\
v_{t}-\Delta_{v}, \psi_{t}-\Delta_{\psi} && v_{t}, \psi_{t}-\Delta_{\psi} && v_{t}+\Delta_{v},   \psi_{t}-\Delta_{\psi}
\end{bmatrix}
.
\end{gather}

\section{Deep Reinforcement Learning Agent}

\subsection{ Partially Observable Markov Decision Process}
The RL problems are usually modeled as Markov Decision Process (MDP). MDP is a tuple $\left\langle S, A, P, R, \gamma  \right\rangle$, where S is the set of states, $A$ is the finite set of actions, $P$ is the transition probability matrix $P_{ss'}^{a} = \mathbb{P}[S_{t+1} = s' | S_t=s,A_t=a ]$, and $R$ is the reward function, where $R_s^a = \mathbb{E}[R_{t+1} | S_t=s, A_t=a]$. $\gamma$ is the discount factor $\gamma \in [0,1] $, which adjusts importance of the future reward.
\par There are two different MDP's based on the observability of the environment: fully and partially observable MDPs. Agent directly observes environment state in full observability. In other words, the state space observed by the agent $S_{t}^{a}$ and the actual state of the environment $S_{t}^{e}$ is same in a given time $t$. On the other hand, in partial observability, the agent indirectly observes the environment state. The state space observed by the agent $S_{t}^{a}$ and the environment $S_{t}^{e}$ are different in a given time $t$. This is named as partially observable markov decision process (POMDP). 
\par The designed aircraft combat environment is the latter, POMDP. Because the state space of the agent is noisy, but there is no noise in the  environment's actual state. This causes the air combat problem even more difficult.

\subsection{Deep Q-Learning}

\par Policy in the RL is a function that returns an action for given state and denoted as $\pi : S \to A $. The main objective of the RL is to find a policy $\pi$ that maximizes cumulative discounted reward at time $t$ as $\mathrm{R}_{t} = \sum_{i=t}^{T}\gamma^{i-t} r_{i}$ where T is terminal state of episode. 
\par The RL algorithms are divided into two sub-branches according to their access to the environment: model-free and model-based. Since the agent has no direct knowledge of state-transition and reward dynamics, we use model-free algorithms in this study. The Deep Q-Learning algorithm (DQN) \cite{2015_nature_DQN}  is a model-free value-based algorithm. It is based on action-value function $Q_{\pi}(s,a) = \mathbb{E}[R_t | s_t= s, a_t=a, \pi]$, which denotes the estimation of return if the agent takes the action $a$ in the state $s$ following the policy $\pi$. The maximum excepted return value that can be reached in a trajectory $\tau$ is denoted as the optimal action-value function $Q^{*}(s,a)$. The DQN algorithm iteratively chooses the maximum action-value at each time step to attain $Q^{*}(s,a)$ using the Bellman equation:

\begin{equation}
\label{equ_bellman_update}
Q_{i+1}(s,a) = \mathbb{E}[r+ \gamma max_{a'} Q_{i}(s',a') | s,a].
\end{equation}

This approach is impractical in experiments since it is difficult to estimate each action-value iteratively. For this reason, The DQN algorithm uses neural networks, which is a non-linear function approximator that can generalize $Q$ values over the state $s$ and action $a$. The neural network takes the states as input, and its output is the $Q$ values. This network can be trained by minimizing loss function $L_{i}(\theta_i)$ at each time step:

\begin{equation}
\label{equ_bellman_update}
y_i = \mathbb{E}_{s'}[r_{i} + \gamma max_{a'} \hat{Q}(s_{i+1},a'; \theta_{i}^{-})],
\end{equation}

\begin{equation}
\label{equ_bellman_update}
L_{i}(\theta_i) = \mathbb{E}[(y_i-Q(s_{i},a_{i};\theta_i))^{2}],
\end{equation}

\noindent where $\theta$ is the network weights. In this study, double DQN \cite{2016_double_dqn} and dueling DQN \cite{2016_dueling_dqn} extensions are combined with vanilla DQN structure \cite{2015_nature_DQN}.

\subsection{Reward Function Design}
Reward function design is an important step in the RL. The designed reward function should maximize the long-term utility. The aim of the air combat is to get behind the opponent aircraft and follow it as much as possible. It is also prevent the enemy agent from following training agent. In previous studies, discrete reward functions are designed \cite{2010_dynamic_programming, 2018_Deep_Q_Ma} according to winning or losing an episode. On the contrary, we use a continouos reward function in this study, and provide its pseudo-code in Algortihm \ref{algo:reward_function}. In this reward function, when the agent gets behind the enemy aircraft, the smaller the angle between the direction of the agent aircraft and the enemy's position, the closer the reward value to 1. On the contrary, it gets closer to -1 when the enemy aircraft is behind us. If there is a crash between two aircraft, the reward is -1.

\begin{algorithm}
\caption{Reward Function}\label{algo:reward_function}
\textbf{Input}: $ATA_a, ATA_e, AA_a, AA_e, R$ = distance
  \begin{algorithmic}[1]
    \State \textbf{if} $(\text{crash}, 0$ m $<=R<10$ m$):$
    \State \hspace{\parindent} $r = -1$
    
    \State \textbf{else if} $(10$ m $<=R<= 100 $ m$):$
    \State \hspace{\parindent} \textbf{if} $(\text{advantage}, |ATA_a|<=30^\circ \quad \text{and} \quad |AA_a|<60^\circ  ):$
    \State \hspace{\parindent}\hspace{\parindent} $r = 1 - \frac{ATA_a}{30} $
    
    \State \hspace{\parindent} \textbf{else if} $(\text{disadvantage}, |ATA_e|<=30^\circ \quad \text{and} \quad |AA_e|<60^\circ  ):$
    \State \hspace{\parindent}\hspace{\parindent} $r = \frac{ATA_e}{30} -1  $
    \State \textbf{else}: 
    \State \hspace{\parindent} $r = 0$
    
  \end{algorithmic} 
\textbf{Output}: $r$ 
\end{algorithm}

\subsection{Noise Reduction with State Stacking}
The perceived position of the enemy aircraft by the agent changes depending on the variance of the noise as shown in Fig. \ref{fig:noisy_state}. The agent can generalize and ignore the noise in low variance cases. However, after the noise exceeds a certain threshold, the agent's performance begins to degrade. 
\par There is a correlation between states, since the positions of the agent are sequential. Therefore, we propose a state stacking technique in order to reduce noise and increase performance. The agent sees the state space at time $t$ is as follows:

\begin{equation}
\label{equ_state_stacking}
S_t = \left\{ S_t, S_{t-1},...,S_{t-n+1} \right\},
\end{equation}
where the $n$ is the stack number. 

\subsection{Enemy Strategy and Self Play}
Another subject emphasized in this study is the strategy of the enemy agent. As stated by Bansal et al. \cite{2018_self_play}, the complexity of the trained RL agent depends on the complexity of the environment. In other words, 
smarter agents can be trained in more complex environments. They also pointed out that self-play can produce substantially more complex behaviors than the environment itself. In addition, self-play also provides many advantages:
\begin{itemize}
    \item Any supervision or human expertise is no longer required.
    \item Agent can learn the environment by playing with itself.
    \item It ensures that the environment difficulty is at the right level to improve agent strategy.
\end{itemize}

\par For the self-play, the enemy agent takes epsilon greedy actions, i.e., the enemy agent chooses the action from the network with probability $1-\lambda$ and take random action from uniform distribution with probabililty $\lambda$. For the specisif case $\lambda = 0$, the enemy agent always use the actions from the network.

\section{Experiments}

The environment is modelled into a closed box as shown in Fig. \ref{fig:noisy_state}, so that the agents would not go beyond a certain distance. During the training phase, the positions and angle values of the agent and the enemy are randomly initialized in the environment. In this way, the agent is forced to explore new states. 
We conducted a hyper-parameter search and the selected hyper-parameter set is given in Table \ref{tab:simulation-hyperparameters}, where we use the same hyper-parameter set for all experiments.

\subsection{State Stacking on Different Noise Levels }

\begin{figure}[hbt!]
    \centering
    \includegraphics[width=1\textwidth]{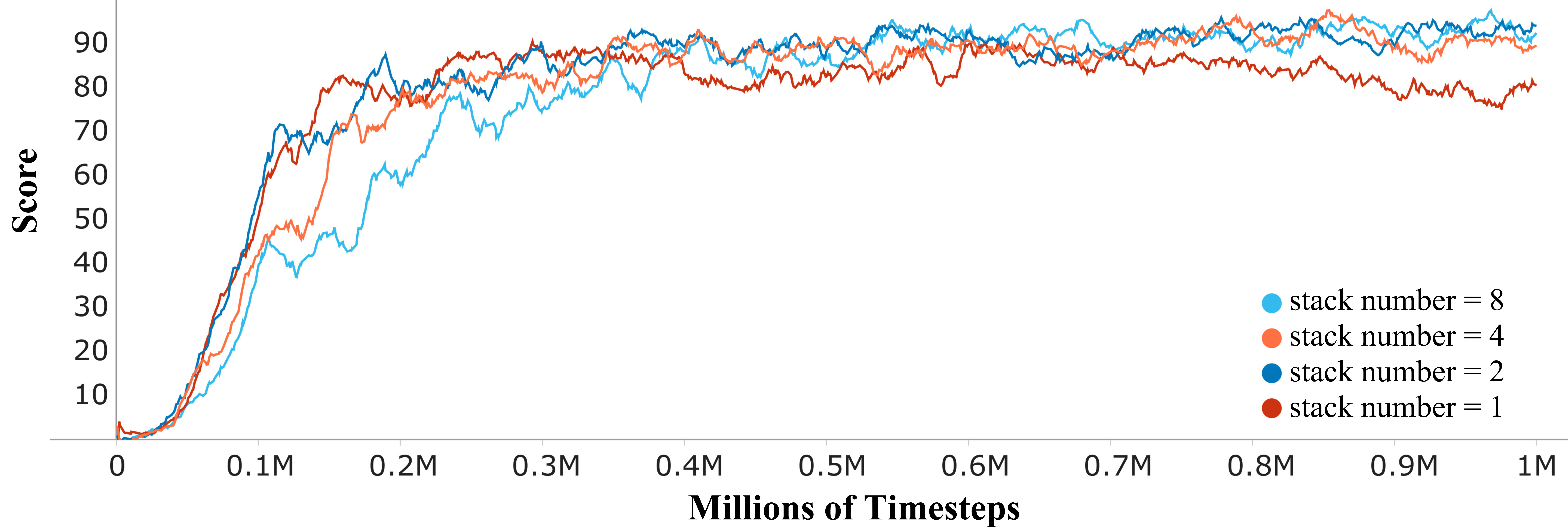}
    \caption{The x-axis shows how many million timesteps the training is, and the y-axis shows the score obtained for a certain moment t. Self-play is not used in this experiment. The position noise value is $n_{x,y} \sim \mathcal{N}(0,5)$ and the angle noise value is $n_{\psi} \sim \mathcal{N}(0,1)$. It can be observed that the performance increases around $\%$10 as the state stack is increased.}
    \label{fig:noise_stack_state_training_5_0}
\end{figure}

\begin{figure}[hbt!]
    \centering
    \includegraphics[width=1\textwidth]{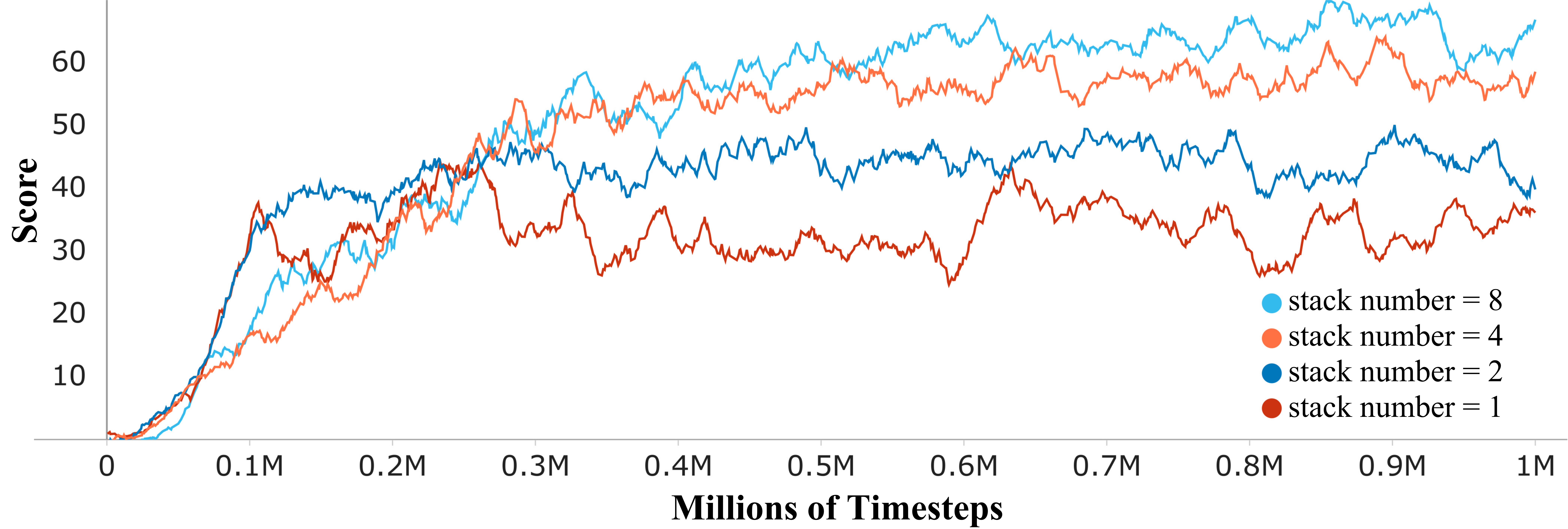}
    \caption{The x-axis shows how many million timesteps the training is, and the y-axis shows the score obtained for a certain moment t. Self-play is not used in this experiment. The position noise value is $n_{x,y} \sim \mathcal{N}(0,20)$ and the angle noise value is $n_{\psi} \sim \mathcal{N}(0,1)$. It can be observed that the performance increases around $\%$59 as the state stack is increased.}
    \label{fig:noise_stack_state_training_20_0}
\end{figure}

Different experiments is carried out on the state stacking approach without using self-play. Two different training graphs with different state noise are given in Fig. \ref{fig:noise_stack_state_training_5_0} and Fig. \ref{fig:noise_stack_state_training_20_0}. In all training processes, the stack number value is chosen as 1, 2, 4, and 8. Firstly, in Fig. \ref{fig:noise_stack_state_training_5_0}, the enemy's position noise value is $n_{x,y} \sim \mathcal{N}(0,5)$ and the enemy's angle noise value is $n_{\psi} \sim \mathcal{N}(0,1)$. In this graph, when the state stack is not used, the reward value is around 90. On the contrary, the agent's performance increased $10\%$ when using any state stack number. In this experiment, i.e., with low noise level, although the use of stacked states provides performance gains, the number of stacks has only a marginal improvement.

 \par Secondly, in Fig. \ref{fig:noise_stack_state_training_20_0}, enemy's position variance is increased to $\sigma_{x,y}=20$, but its angle variance is remained as $\sigma_{\psi}=1$. In this case, the environment is quite noisy, and the episode score is around 44 without using the stack. When the number of state stack is 8, the score increases to around 70. In other words, $\%$59 performance increase is achieved compared to the situation where no stack is used. In addition, it is clearly shown in this graph that the score increases in proportion to the number of stacks. This gives information about how robust the proposed method is.

\par Furthermore, training is performed with different combinations of state stack and noise values as shown in Table \ref{tab:noise_stack}. This table shows that as the noise value increases, obtained highest score decreases inversely. Stack number does not affect the performance where the position's noise variance is low ($\sigma^2_{x,y} = 1$ and $\sigma^2_{x,y} = 3$). However, when noise exceeds these values, stacking starts to compensate the noise.

\begin{table}[hbt!]
\caption{\label{tab:noise_stack} Highest scores in different variation of noise and state stack values. Normalized stacking scores is calculated as: 100 x (highest stacking score / \# of stack=1 ) }
\centering
\begin{tabular}{lccccc}
\hline
Environment noise & $\#$ of stack = 1 & $\#$ of stack = 2& $\#$ of stack = 4&$\#$ of stack = 8   & Normalized stacking score\\\hline
$n_{x,y} \sim \mathcal{N}(0,1)$  &\textbf{107.3} & 105.1	&106.9&	104.6 & \% 99,62  \\
$n_{x,y} \sim \mathcal{N}(0,3)$& \textbf{102.8}	&102.1&	101.6 &	\textbf{102.8} & \%100\\
$n_{x,y} \sim \mathcal{N}(0,5)$& 90.82	&96.11&	97.65&	\textbf{98.25} &  \%108.18\\
$n_{x,y} \sim \mathcal{N}(0,10)$& 68.2&80.92&	85.79&	\textbf{88.89} &  \%130.33\\
$n_{x,y} \sim \mathcal{N}(0,20)$& 44.17&50.11&	64.1&\textbf{70.47} & \%159.54\\
$n_{x,y} \sim \mathcal{N}(0,40)$& 19.11&	34.39&	39.95&	\textbf{45.49} & \%238.04 \\
\hline
\end{tabular}
\end{table}



\subsection{Training with Self-play}
There are two sides in the air combat problem. Better strategies can emerge if competition is created between these sides during training. One of the purpose of the self-play is to employ this contest. In this context, we design a training process that includes self-play. In this experiment setup, enemy policy $\pi_e^{t}$ is replaced with agent policy $\pi_a^{t}$ at every $n$ timesteps. The parameter $n$ is selected as 50.000 by using grid search. In addition, the enemy is forced to choose random action with  the probability $\lambda$ and action from its own policy with the probability $1-\lambda$.

\par
In this experiment, we provide the performance of self-play agent with different exploration parameter $\lambda$. In the evaluation, the enemy agent is the one, which is trained without self-play and the result are shown in Table \ref{tab:self_play_comparison}.  

Both the training agents and the enemy agent are trained in the same environment conditions, where the noise variances are $\sigma_{x,y} = 10 \text{ and } \sigma_{\psi} = 1$.
These two agents also did not encounter each other during the training.
At the beginning of the simulation, both agents are initialized at random positions and angles.
This experiment is evaluated by using 1000-run Monte Carlo simulation.
Win, lose and tie conditions are designed according to the reward function.
When any agent achieves the strike angle within striking distance, it is the winner.
The maximum episode length is set as 200 timesteps.
If both agents are alive at the end of the episode, then tie condition occurs. 

\par As seen in Table \ref{tab:self_play_comparison}, it is obvious that self-play agents outperformed each match regardless of the parameter $\lambda$.
However, for the lower values of the parameter $\lambda$, which correspond to the more sophisticated enemy agents during training, the training agent achieved better performances in the test stage.

\begin{table}[hbt!]
\caption{\label{tab:self_play_comparison} Agents trained with self-play and without self-play launched from random locations and faced 1000 times. Noise value's variances are $\sigma_{x,y} = 10 \text{ and } \sigma_{\psi} = 1$. The win probability is calculated as: win/(win+lose).}
\centering
\begin{tabular}{cccccc}
\hline
      Agent & Enemy & Win & Lose & Tie & Win Probability\\\hline
Self-play $\lambda=0$ &  $n_{x,y} \sim \mathcal{N}(0,10)$ &726 &	99	&175	&0.88\\
Self-play $\lambda=0.2$&  $n_{x,y} \sim \mathcal{N}(0,10)$ &753 &	102&	145& 0.88\\
Self-play $\lambda=0.5$&  $n_{x,y} \sim \mathcal{N}(0,10)$ &758 &	109&	133&	0.87\\
Self-play $\lambda=0.9$&  $n_{x,y} \sim \mathcal{N}(0,10)$ &574 &	309&	117&	0.65\\

\hline
\end{tabular}
\end{table}

\section{Conclusion}
We developed an air combat simulator, which provide the agents with noisy observations. 
The inclusion of noise greatly reduced the performance of the agents using a single observation.
On the other hand, the agents using stacked observations achieve higher scores.
We compared the performance of the agents using different number of stacked states at different noise levels. 
We demonstrated that the performance increases as number of stacks rises in noisy environments.
The stacked states contain temporal information and since the consecutive observations are highly related, it inherently leads to the noise reduction.
In addition, we constructed a self-play architecture in competitive setup, which provides smarter enemy agent strategies during the training phase.
We empirically showed that the self-play agents obtain better results compared to the non self-play agents in noisy environments. 

\par As future work, the stacking structure can be compared by implementing different RL algorithms. The curriculum learning setup can be use for noisy air combat environment in order to increase performance because noise level can be increased gradually so that the agent makes a smooth transition from easy environment to difficult. 

\section*{Appendix}

\begin{table}[hbt!]
\caption{\label{tab:simulation-hyperparameters} Simulation Hyper-parameters }
\centering
\begin{tabular}{lc}
\hline \hline
State space size&7\\
Action space size& 9	\\
Learning rate, $lr$ & 1e-4\\
Discount factor, $\gamma$ & 0.99\\
Replay memory size & 20.000\\
Mini batch size& 32\\
Target update frequency, C & 10.000\\
Initial exploration rate & 1\\
Final exploration rate& 0.05\\
Final exploration timesteps & 100.000\\
Episode length & 200 \\

\hline
\end{tabular}
\end{table}

\bibliography{sample}

\end{document}